\documentclass{article}
\usepackage{spconf,amsmath,graphicx,hyperref}
\usepackage{float}
\usepackage{amsfonts}

\title{EmoHeal: An End-to-End System for Personalized Therapeutic Music Retrieval from Fine-Grained Emotions}
\name{Xinchen Wan, Jinhua Liang, Huan Zhang}
\address{Queen Mary University of London, London, UK}
\begin{document}
\ninept
\maketitle
\begin{abstract}
Existing digital mental wellness tools often overlook the nuanced emotional states underlying everyday challenges. For example, pre-sleep anxiety affects more than 1.5 billion people worldwide, yet current approaches remain largely static and ``one-size-fits-all'', failing to adapt to individual needs. In this work, we present EmoHeal, an end-to-end system that delivers personalized, three-stage supportive narratives. EmoHeal detects 27 fine-grained emotions from user text with a fine-tuned XLM-RoBERTa model, maping them to musical parameters via a knowledge graph grounded in music therapy principles (GEMS, iso-principle). EmoHeal retrieves audiovisual content using the CLaMP3 model to guide users from their current state toward a calmer one (``match-guide-target''). A within-subjects study (N=40) demonstrated significant supportive effects, with participants reporting substantial mood improvement (M=4.12, $p<0.001$) and high perceived emotion recognition accuracy (M=4.05, $p<0.001$). A strong correlation between perceived accuracy and therapeutic outcome ($r=0.72$, $p<0.001$) validates our fine-grained approach. These findings establishes the viability of theory-driven, emotion-aware digital wellness tools and provides a scalable AI blueprint for operationalizing music therapy principles.
\end{abstract}
\begin{keywords}
Digital Wellness, Emotion Recognition, Music Healing, Multimodal Content Retrieval, Personalized Intervention
\end{keywords}
\section{Introduction}
\label{sec:intro}
The need for accessible mental wellness tools is critical, with anxiety disorders affecting 359 million people globally \cite{who2021anxiety} and the digital mental health market projected to exceed \$27 billion by 2025 \cite{researchandmarkets2025digital}. However, dominant applications like Calm\footnote{https://www.calm.com/} and Headspace\footnote{https://www.headspace.com/} rely on a static, library-based paradigm, requiring users to self-diagnose and manually select pre-recorded content. This model suffers from two fundamental flaws: a \textbf{static approach} that fails to adapt to users' free-form expression, and a \textbf{lack of therapeutic dynamism} required to implement principles like the ``iso-principle,'' which often creates an ``emotional mismatch.''

To address these limitations, we propose EmoHeal, a novel system that creates personalized supportive journeys by guiding users through a three-stage ``match-guide-target'' narrative that operationalizes the iso-principle. It integrates three core components: a fine-tuned XLM-RoBERTa model for fine-grained emotion recognition, a knowledge graph to map emotions to musical parameters, and a CLaMP3 model for real-time content retrieval. Our contributions are threefold: (1) a novel end-to-end system integrating these state-of-the-art AI technologies; (2) a computational blueprint for operationalizing music therapy principles; and (3) an empirical validation through a 40-participant user study demonstrating the efficacy of our theory-driven, personalized approach.

\begin{figure*}[thb]
    \centering
    \centerline{\includegraphics[width=\textwidth]{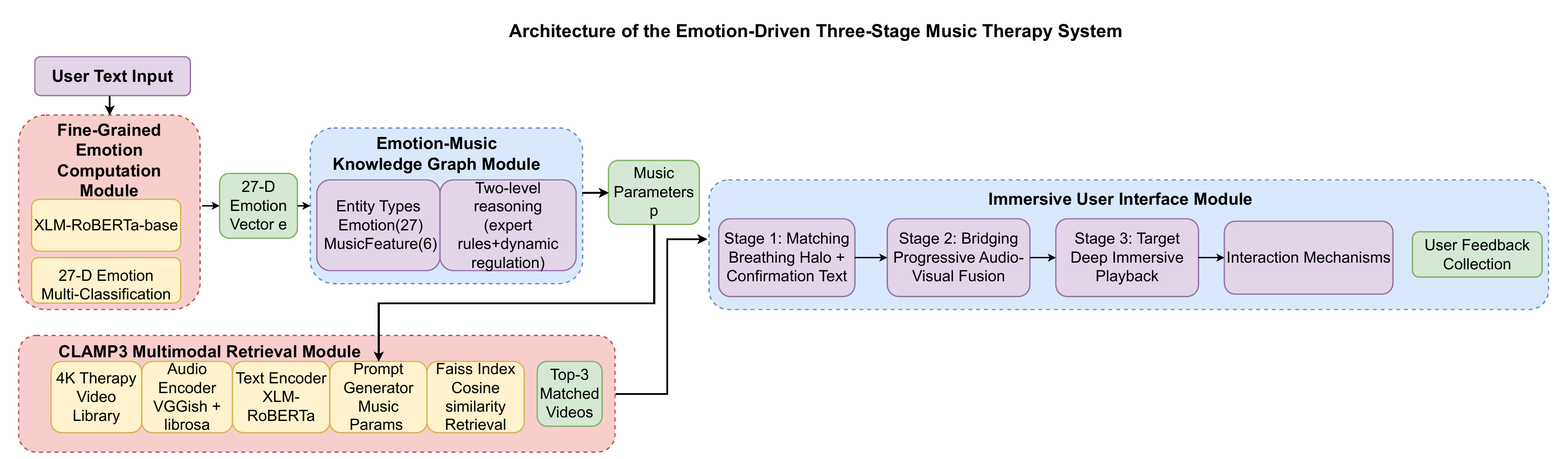}}
    \caption{Architecture of the Emotion-Driven Three-Stage Music Therapy System. The system integrates four core modules: a Fine-Grained Emotion Computation module to analyze user text, an Emotion-Music Knowledge Graph to translate emotions into musical parameters, a CLaMP3 Multimodal Retrieval Module to find corresponding audiovisual content, and an Immersive User Interface to present the therapeutic journey.}
    \label{fig:sys_arc}
\end{figure*}

\section{Related Work}
\label{sec:related}
Our work integrates advancements from four key domains. First, in fine-grained emotion recognition, some advanced approaches such as XLM-RoBERTa \cite{conneau2020unsupervised,jia2024bridging} have provided powerful language representations capable of capturing subtle affective cues. Together with large-scale benchmarks like GoEmotions \cite{demszky2020goemotions}, these developments have enabled a shift beyond coarse sentiment analysis toward more nuanced emotion understanding. These models have proven effective in complex domains, including health-related online contexts \cite{khanpour2018fine}. However, a key challenge remains in applying these models, often trained on public social media data, to the more private and nuanced characteristic of wellness setting \cite{acheampong2020textbased}.

The second pillar is \textbf{computational music therapy}. Our work is grounded in psychological frameworks like GEMS, which defines music-specific emotional dimensions \cite{zentner2008emotions}, and long-standing therapeutic theories such as the iso-principle \cite{hanser1999handbook}. Neuroscience research, showing that music engages unique emotional processing pathways in the brain \cite{koelsch2006investigating}, further motivates music's use as a therapeutic modality. The primary challenge here is computationally operationalizing these complex, dynamic principles into an automated system.

For content delivery, we build on advances in multimodal retrieval. Early audio–language models such as MuLan and CLAP established strong text–audio alignment~\cite{wu2022mulan,huang2022clap}, while some recent systems like CLaMP3~\cite{wu2025clamp3,liang2023adapting} further improved scalability and retrieval accuracy. More recently, large language models (LLMs) have been extended to audio~\cite{liang2023acoustic,10890522}, showing promise as general-purpose learners, but their heavy computational demands restrict deployment in resource-constrained settings. In parallel, generative models such as MusicLM can synthesize novel music~\cite{agostinelli2023musiclm,hisariya2025bridging,zhang2024dexter,zhang2025renderboxexpressiveperformancerendering,yuan2023leveraging}, though their potential in wellness applications remains largely untapped. LLM-driven audio generators~\cite{liu2025wavjourney,liang2024wavcraft} adapt to task-specific needs by integrating specialist models, yet they face the same efficiency challenges as other LLM-based systems. In contrast, our retrieval-based strategy ensures high-fidelity, professionally produced audiovisual content while remaining computationally efficient—an essential requirement for wellness applications.

Finally, the user experience is informed by \textbf{Human-Computer Interaction (HCI)}. We apply concepts of ``Calm Technology'' \cite{weiser1996calm} and established guidelines for digital wellness tools \cite{kvedar2022digital}. Furthermore, creating a sense of ``presence'' and immersion is known to enhance the efficacy of virtual environments \cite{slater2003note}, a principle we aimed to incorporate. A key research gap remains in designing for users in vulnerable emotional states, where traditional usability heuristics may not apply.

While progress in each area is significant, prior work has often been fragmented. Few systems cohesively merge these technical and psychological pillars into a single, end-to-end pipeline evaluated on user-centric outcomes. EmoHeal directly addresses this integration gap.

\begin{figure}[htb]
    \centering
    \centerline{\includegraphics[width=0.5\columnwidth]{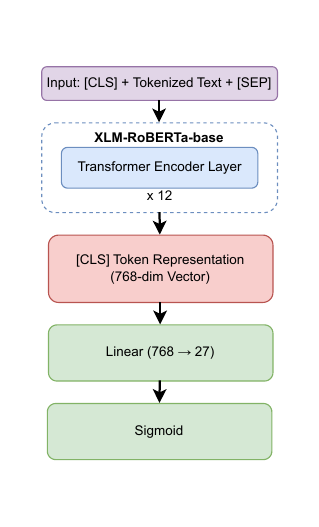}}
    \caption{Architecture of the Emotion Computation Module.}
    \label{fig:ac_module}
\end{figure}

\section{THE PROPOSED EMOHEAL SYSTEM}
\label{sec:method}
As illustrated in Fig.~\ref{fig:sys_arc}, our system, EmoHeal, is an end-to-end pipeline that generates personalized supportive music experiences by processing user text input into a final audio-visual presentation.

\begin{figure*}[thb]
    \centering
    \centerline{\includegraphics[width=0.8\textwidth]{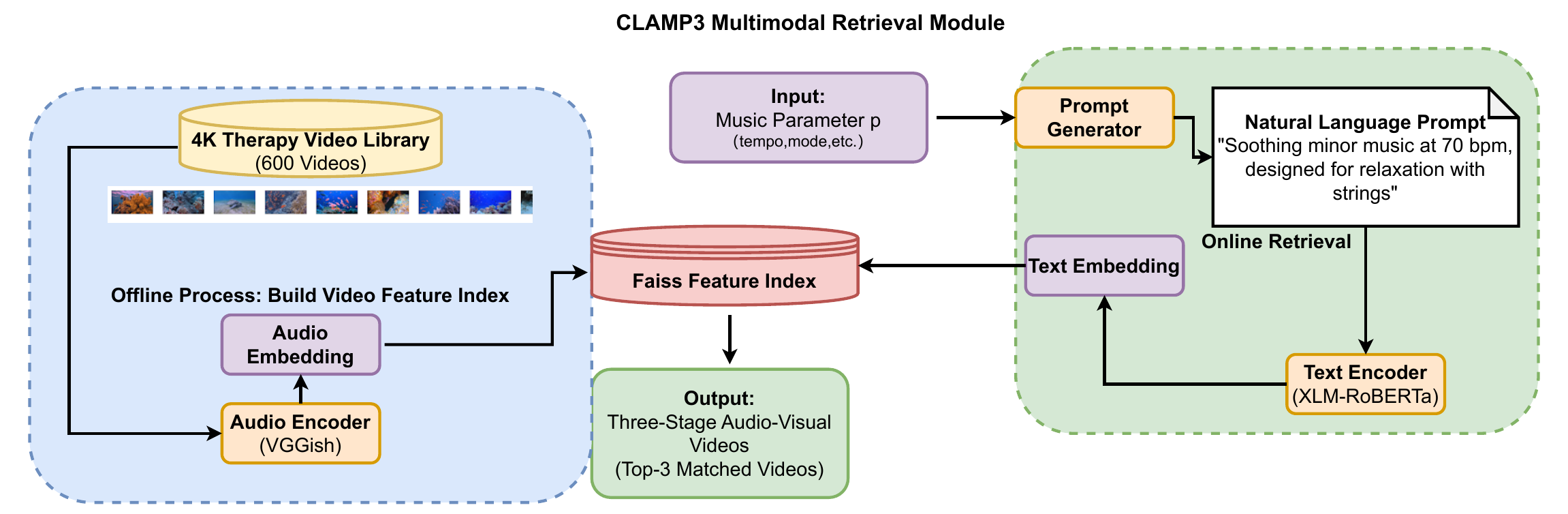}}
    \caption{Workflow of the CLaMP3 Multimodal Retrieval Module.}
    \label{fig:mi_module}
\end{figure*}

\subsection{Fine-Grained Emotion Computation}
\label{ssec:emotion_module}
To capture nuanced emotional states, we selected XLM-RoBERTa-base as our backbone network. As shown in Fig.~\ref{fig:ac_module}, a linear classification head ($768 \times 27$) is added on top of the model's [CLS] token representation to predict a 27-dimensional probability vector, $e \in \mathbb{R}^{27}$, corresponding to the 27 fine-grained emotion classes defined by the GoEmotions dataset \cite{demszky2020goemotions}. For this task, the pre-trained XLM-ROBERTa-base model was fine-tuned on a custom multi-corpus dataset. This dataset combines two sources: the foundational English GoEmotions dataset \cite{demszky2020goemotions}, which defines our 27 fine-grained emotion labels, and the NLPCC-2014 Chinese Emotion Analysis Dataset. To align the label spaces, as the NLPCC-2014 dataset uses a different set of coarser emotion labels, we developed a programmatic mapping strategy. For instance, the coarse label `joy' from the Chinese dataset was heuristically mapped to a multi-label vector where the corresponding fine-grained GoEmotions labels for `joy', `amusement', and `excitement' were all activated (set to 1), while all other labels were set to 0. The training utilized the Focal Loss function, formulated as:
\begin{equation}
\label{eq:focal_loss}
L_{cls} = -\alpha_{t}(1-p_{t})^{\gamma}\log(p_{t})
\end{equation}
where $p_t$ is the class probability and we set the focusing parameter $\gamma=2$. The model was fine-tuned using an AdamW optimizer (LR=2e-5) with 5-fold cross-validation, achieving a validation Macro-F1 of 0.64 and Weighted-F1 of 0.71.

\begin{figure}[thb]
    \centering
    \centerline{\includegraphics[width=0.9\columnwidth]{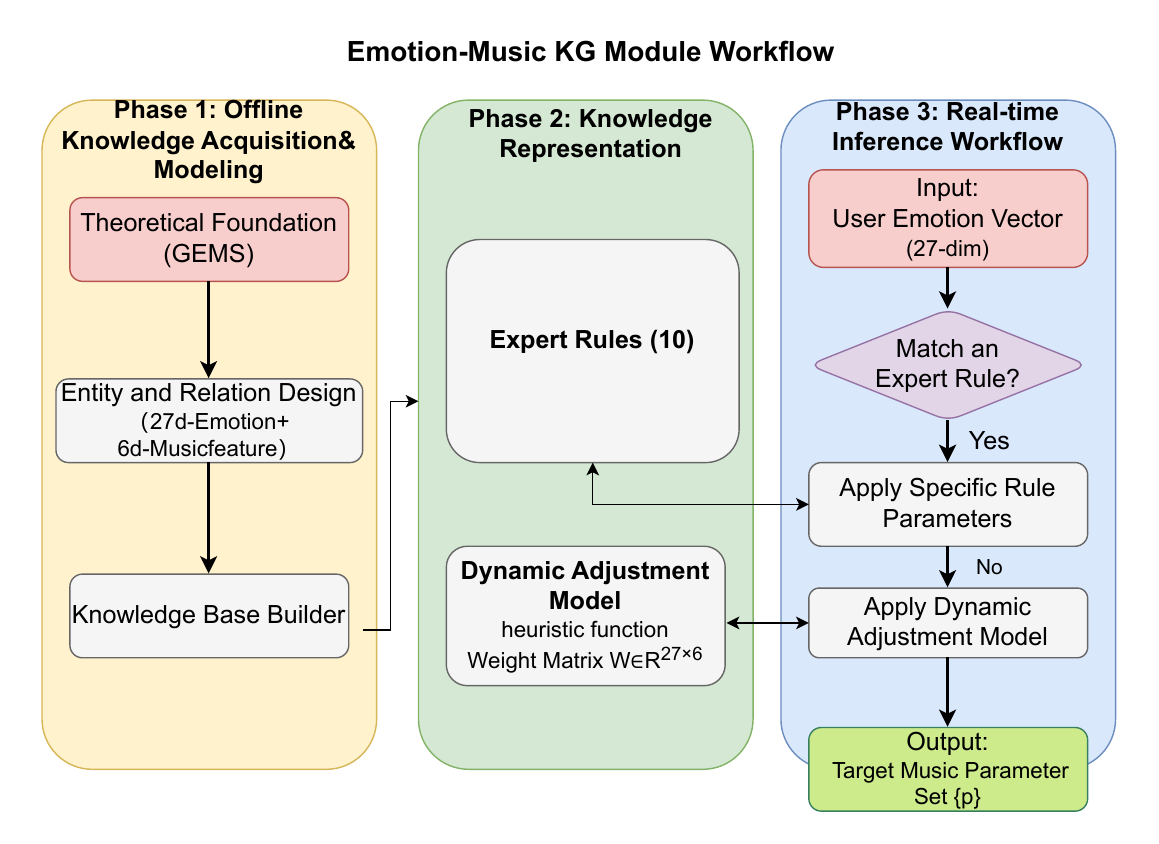}}
    \caption{Workflow of the Emotion-Music KG Module.}
    \label{fig:kg_module}
\end{figure}

\subsection{Emotion-Music Knowledge Graph}
\label{ssec:kg_module}
This module translates the emotion vector $e$ into a set of six musical parameters $p$, including tempo (continuous, e.g., 60-120 BPM), mode (e.g., major/minor), timbre (e.g., bright/dark), harmony (e.g., consonant/dissonant), register (e.g., high/low), and density (e.g., sparse/dense). The logic, grounded in music psychology \cite{zentner2008emotions}, is operationalized via a two-tier inference system shown in Fig.~\ref{fig:kg_module}. The first tier handles unambiguous emotions, defined as cases where a single primary emotion score exceeds a high-intensity threshold, which was set empirically to $\tau=0.7$ based on pilot testing. For these cases, an expert-curated rule is triggered to provide a rapid, consistent response, following the form:
\[
\text{IF } e_i > \tau \ \text{ THEN } \ p_j = v,
\]
where $e_i$ is the score for an emotion, $p_j$ a musical parameter, and $v$ a target value, directly applying the aforementioned threshold $\tau$.

If no rule is triggered, the vector is treated as a complex emotion (e.g., a mix of `nostalgia' and `optimism') and passed to the second tier: a dynamic adjustment model. This model projects the vector $e$ through a handcrafted, theory-driven weight matrix $W \in \mathbb{R}^{27 \times 6}$. This matrix, derived from music emotion recognition literature \cite{panda2020audio}, contains normalized weights in [-1.0, 1.0] encoding the influence of each emotion on each parameter. The final blended parameters are yielded by:
\[
p = \sigma(e W),
\]
where $p \in \mathbb{R}^{6}$ represents the continuous parameter set and $\sigma(\cdot)$ is a normalization function. This weighted mapping enables a fine-grained adaptation beyond categorical rules.

\begin{figure*}[t]
    \centering
    \centerline{\includegraphics[width=0.95\textwidth]{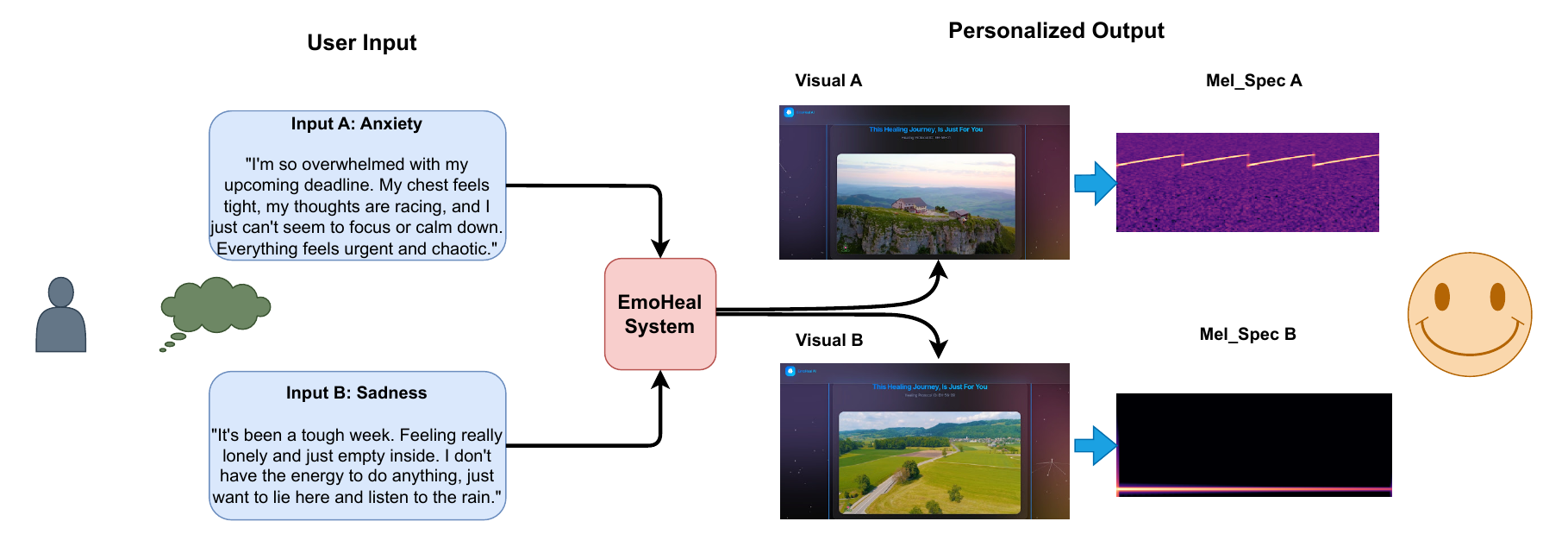}} 
    \caption{Demonstration of the EmoHeal system. Two distinct user inputs describing anxiety (top) and sadness (bottom) are processed by the system, resulting in distinct, personalized audiovisual outputs. The audio component is visualized as a mel spectrogram to illustrate the difference in acoustic features.}
    \label{fig:demo}
\end{figure*}

\subsection{Multimodal Content Retrieval}
\label{ssec:retrieval_module}
A key part of our methodology was the construction of a custom library of over 600 3-minute 4K video clips. These were derived from commercially licensed long-form films from `Nature Relaxation', curated based on principles from environmental psychology, particularly Attention Restoration Theory (ART) \cite{kaplan1995restorative}. A programmatic pipeline segmented these films by first detecting scene boundaries via color histograms, then identifying 'calm segments' with low motion magnitude using optical flow, and finally partitioning these segments into non-overlapping 3-minute clips. For offline feature extraction, each clip's audio track was converted into a fixed-dimension 128-d embedding using a pre-trained VGGish model \cite{hershey2017cnn} and temporal average pooling. These embeddings were indexed into a Faiss library using an Inverted File (IVF) index to enable near real-time Approximate Nearest Neighbor (ANN) search.

The real-time retrieval process is enabled by the CLaMP3 model \cite{wu2025clamp3}, whose jointly trained text and audio encoders map semantically similar content to nearby vectors in a shared multimodal embedding space, thus allowing for direct text-to-audio similarity comparison. During a user session, as detailed in Fig.~\ref{fig:mi_module}, the musical parameters $p$ from the knowledge graph are first converted into a descriptive natural language prompt using a template-based function. This prompt is then fed into the CLaMP3 text encoder to generate a query text embedding. This query is used to perform a cosine similarity search against the pre-indexed audio embeddings, and the top-3 most similar videos are retrieved for the user.

\subsection{User Interface and Interaction Design}
The system was implemented as a web application\footnote{A project showcase, including video demonstrations, is available at: [https://jeweled-scarer-812.notion.site/EmoHeal-Project-Showcase-27278a073579807b8e82e5fac88b821f]}. The front-end, designed for a responsive and dynamic user experience, was built with HTML5, CSS3, and JavaScript (ES6+), utilizing the Bootstrap framework for its UI components. The back-end consists of a Python-based REST API, developed with the Flask framework, which serves the core machine learning models. The interface design follows "Calm Technology" principles \cite{weiser1996calm}, employing a dark theme and a "progressive disclosure" model to minimize cognitive load and protect user privacy by using all data ephemerally.

\section{EVALUATION}
\label{sec:eval}
\subsection{Experimental Setup}
A total of 40 participants were recruited for the within-subjects study. The cohort consisted primarily of students from Queen Mary University of London, recruited using a snowball sampling method through university-affiliated WhatsApp and WeChat student groups. The demographics of the participants are as follows: the mean age was 26.2 years (SD=4.8), with a gender distribution of 22 female and 18 male. Regarding language proficiency, 60.0\% reported English as their primary language, while 40.0\% reported Chinese. As this was a non-clinical study, inclusion criteria required participants to have no severe mental health conditions requiring clinical intervention.

The experimental procedure was straightforward. Each participant provided a text description of their current mood, watched the 3-minute therapeutic video generated in real-time by the system, and subsequently completed a post-session questionnaire. Primary outcomes were assessed using a 5-point Likert scale (1=Strongly Disagree, 5=Strongly Agree), where participants rated their agreement with statements such as, ``The generated video accurately reflected the emotions I described in my text'' (System Responsiveness) and ``Watching the video had a positive impact on my mood'' (Supportive Effect). We also collected qualitative feedback through open-ended questions.

\subsection{Results}
To provide a qualitative illustration of the system's personalization capabilities, Fig.~\ref{fig:demo} showcases how two distinct emotional inputs result in markedly different audiovisual outputs. The primary quantitative results, summarized in Table~\ref{tab:descriptives} and Table~\ref{tab:correlation}, confirm the system's effectiveness. It demonstrated a significant positive impact on user mood (M=4.12, $p<0.001$) and was perceived as highly responsive to their described emotions (M=4.05, $p<0.001$). The high ratings for overall atmosphere and multimodal coherence (both M=4.18) further suggest a high-quality user experience. These findings are further supported by the rating distributions, where 85.0\% of participants rated the emotion accuracy and 87.5\% rated their mood improvement as ``Agree'' (Score 4) or ``Strongly Agree'' (Score 5). A notable 78\% also explicitly mentioned feeling ``understood'' by the system's personalized response.

Our key finding, detailed in Table~\ref{tab:correlation}, is the strong, positive correlation between the perceived accuracy of emotion recognition and the subsequent mood improvement ($r=0.72, p<0.001$). This statistically validates our entire fine-grained approach, providing the empirical backbone for the following discussion.

\begin{table}[t]
\caption{Primary Outcome Measures - Descriptive Statistics (N=40). All results are significant ($p<0.001$) via one-sample t-tests against a neutral midpoint of 3.0.}
\label{tab:descriptives}
\centering
\begin{tabular}{p{4 cm} c c}
\hline
\textbf{Measure} & \textbf{Mean (M)} & \textbf{Std. Dev. (SD)} \\
\hline
\textbf{RQ1: Healing Effects} & & \\
Positive mood impact & 4.12 & 0.89 \\
\hline
\textbf{RQ2: System Responsiveness} & & \\
Perceived emotion accuracy & 4.05 & 0.83 \\
\hline
\textbf{RQ3: User Experience} & & \\
Overall atmosphere & 4.18 & 0.76 \\
\hline
\textbf{RQ4: Content Quality} & & \\
Multimodal coherence & 4.18 & 0.76 \\
\hline
\end{tabular}
\end{table}

\begin{table}[t]
\caption{Correlation Analysis.}
\label{tab:correlation}
\centering
\begin{tabular}{p{5.5cm} c c}
\hline
\textbf{Variable Pair} & \textbf{r} & \textbf{p-value} \\
\hline
Emotion accuracy $\times$ Mood improvement & 0.72 & $<0.001$ \\
Text length $\times$ Emotion accuracy & 0.31 & $<0.05$ \\
\hline
\end{tabular}
\end{table}

\section{Discussion and Conclusion}
\label{sec:discussion_conclusion}

This work introduced \textit{EmoHeal}, an AI system that translates fine-grained emotional states from text into personalized, theory-driven musical experiences. Results show that moving beyond coarse emotion models is essential: recognition accuracy strongly correlated with therapeutic efficacy ($r=0.72$), underscoring the role of nuanced recognition in fostering a sense of being ``understood.'' 

A key contribution is the hybrid design that combines neural networks with a knowledge graph grounded in constructs such as GEMS and the iso-principle. This neuro-symbolic approach produced coherent and interpretable mappings, enhancing both theoretical grounding and trust. Participants also valued the system’s simplicity and personalization, which reduced cognitive load and supported emotional processing. Limitations include the single-session design, reliance on self-report, and a homogenous sample. Future work should extend to diverse populations, multimodal signals, and clinical integration. 

Overall, \textit{EmoHeal} demonstrates that fine-grained emotion recognition coupled with hybrid AI can deliver measurable therapeutic benefits and provide a scalable blueprint for digital wellness tools.


\vfill\pagebreak

\bibliographystyle{IEEEbib}
\bibliography{strings,refs}

\end{document}